\documentclass{article}
\usepackage{ijcai17}

\usepackage{amsmath, amssymb}
\usepackage{type1cm}

\usepackage{bm}
\usepackage{url} 
\usepackage{times}
\usepackage{lipsum}
\usepackage{xcolor}
\usepackage{comment}
\usepackage{amsmath}
\usepackage[font={small},textfont={small}]{caption}
\usepackage{wrapfig}
\usepackage{multirow}
\usepackage{graphicx}
\usepackage{booktabs}
\usepackage{graphicx}
\usepackage{mathtools}
\usepackage{subfigure}
\usepackage{algorithm}
\usepackage{algorithmic}

\newcommand\AuthorMark[1]{\textsuperscript{#1}}

\newcommand{\gold}{\text{gold}}
\newcommand{\aux}{\text{aux}}

\newcommand{\KB}{\mathcal{G}}
\newcommand{\head}{\text{head}}
\newcommand{\tail}{\text{tail}}

\newcommand{\ul}[1]{\underline{#1}}

\usepackage[hidelinks]{hyperref}

\begin{document}
	\title{Knowledge Transfer for Out-of-Knowledge-Base Entities:\\A Graph Neural Network Approach}

	\author{
		Takuo Hamaguchi{\rm,}\AuthorMark{1}\enskip
		Hidekazu Oiwa{\rm,}\AuthorMark{2}\enskip
		Masashi Shimbo{\rm,}\AuthorMark{1}\and
		Yuji Matsumoto\AuthorMark{1} \\
		\AuthorMark{1}Nara Institute of Science and Technology, Ikoma, Nara, Japan \\
		\AuthorMark{2}Recruit Institute of Technology \\
		\{\href{mailto:takuo-h@is.naist.jp}{takuo-h},\href{mailto:shimbo@is.naist.jp}{shimbo},\href{mailto:matsu@is.naist.jp}{matsu}\}@is.naist.jp,
		\href{mailto:oiwa@recruit.ai}{oiwa@recruit.ai}
	}

	\maketitle
	\begin{abstract}
	Knowledge base completion (KBC) aims to predict missing information in a knowledge base.
	In this paper, we address the out-of-knowledge-base (OOKB) entity problem in KBC:
	how to answer queries concerning test entities not observed at training time.
	Existing embedding-based KBC models assume that all test entities are available at training time,
		making it unclear how to obtain embeddings for new entities without costly retraining.
	To solve the OOKB entity problem without retraining,
		we use graph neural networks (Graph-NNs)
		to compute the embeddings of OOKB entities,
		exploiting the limited auxiliary knowledge provided at test time.
	The experimental results show the effectiveness of our proposed model in the OOKB setting.
	Additionally, in the standard KBC setting in which OOKB entities are not involved,
		our model achieves state-of-the-art performance on the WordNet dataset.
	The code and dataset are available at \protect{\url{https://github.com/takuo-h/GNN-for-OOKB}}.
\end{abstract}


	\section{Introduction}
	{\em Knowledge bases} such as WordNet \cite{Wordnet} and Freebase \cite{Freebase}
		are used for many applications
		including information extraction, question answering, and text understanding.
	These knowledge bases can be viewed as a set of \emph{relation triplets},
		i.e., triplets of the form $(h, r, t)$
		with an entity $h$ called the \emph{head entity},
		a relation $r$,
		and an entity $t$ called the \emph{tail entity}
		\cite{transE,neig}.
	Some examples of relation triplets are
		({\it Philip-K.-Dick}, {\it write}, {\it Do-Androids-Dream-of-Electric-Sheep?})
		and ({\it Do-Androids-Dream-of-Electric-Sheep?}, {\it is-a}, {\it Science-fiction}).
	Although a knowledge base contains millions of such triplets,
		it is known to suffer from incompleteness \cite{review}.
	Knowledge base completion (KBC) thus aims to predict the information missing in knowledge bases.

	In recent years, \emph{embedding-based} KBC models have been successfully
		applied to large-scale knowledge bases\cite{transH,transR,transD,transG,neig,trav}.
	These models build the distributed representations (or, \emph{vector embeddings}) of entities and relations
		observed in the training data, and use various vector operations
		over the embeddings to predict missing relation triplets.

\begin{figure}[!t]
	\centering
	\includegraphics[width=.75\linewidth]{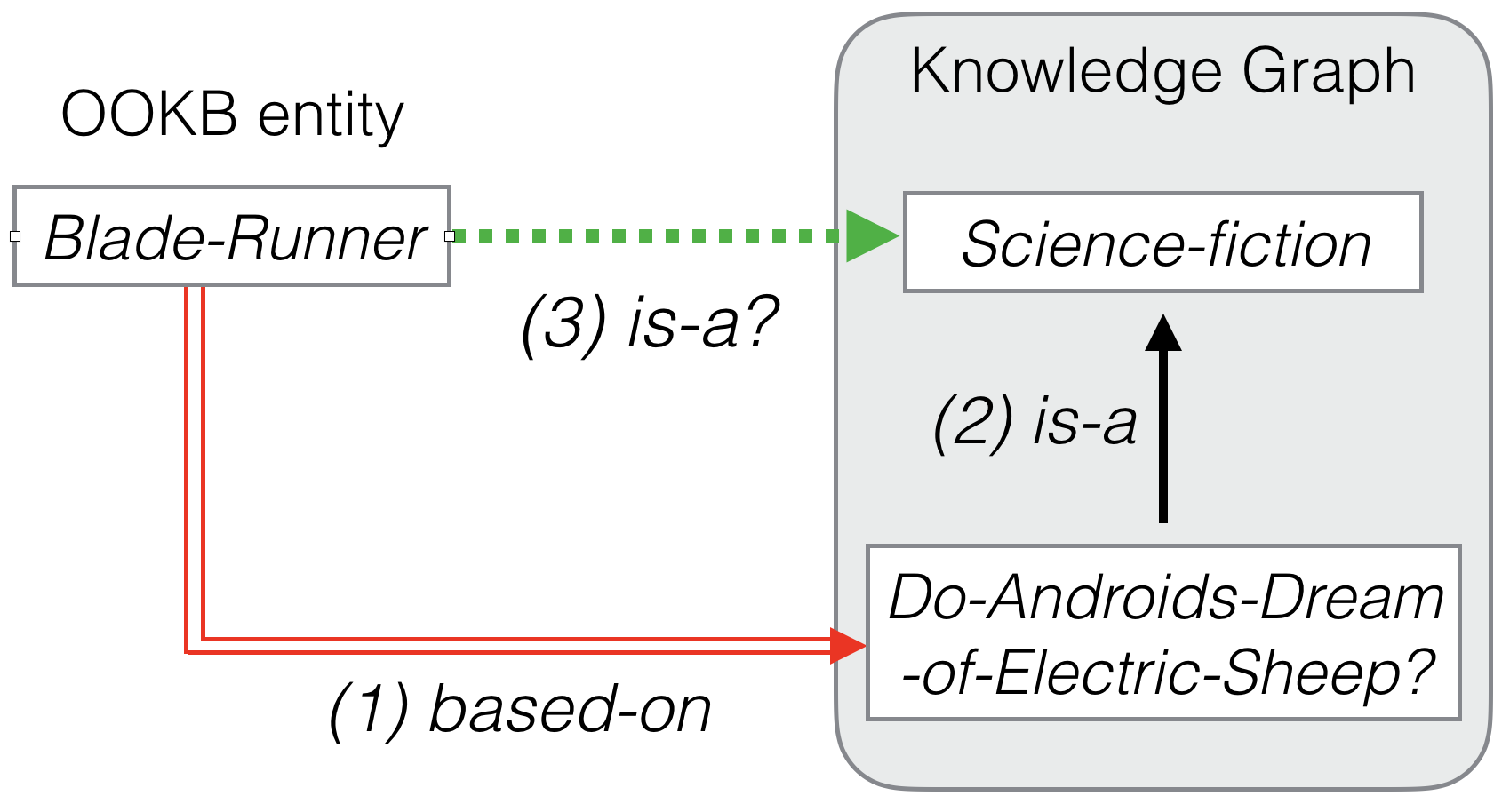}
	\caption{
		OOKB entity problem.
		At test time, we receive a new triplet (1) depicted as a red double arrow,
			which contains an entity ``{\it Blade Runner}'' that is not observed
			in the knowledge graph (shown as the shaded box).
		The task is to tell whether any other relations hold between \emph{Blade Runner}
			and the entities in the knowledge graph,
			by using (1) the new triplet and (2) the existing triplet, depicted by a black arrow.
		For example, we want to answer the question, ``Is Blade Runner science fiction?'',
			i.e., whether (3) the green dashed arrow in the figure should be drawn.
	}
	\label{image:relation-example}
\end{figure}

	In this paper, we address the {\em out-of-knowledge-base} (OOKB) entity problem in embedding-based KBC.
	This problem arises when new entities (OOKB entities) occur in the relation triplets that are given to the system \emph{after} training.
	As these entities were unknown to the system at training time, the system does not have their embeddings,
		and hence does not have a means to predict the relations for these entities.
	The OOKB entity problem thus asks how to perform KBC involving such OOKB entities.
	Although it can be solved by retraining the embeddings using the added relation triplets containing the OOKB entities,
		a solution avoiding costly retraining is desirable.

	This problem is of practical importance because OOKB entities crop up whenever new entities,
		such as events and products, are produced, which happens everyday.
	For example,
	suppose we find an OOKB entity ``{\it Blade-Runner}''
		in a new triplet
		({\it Blade-Runner}, {\it based-on}, {\it Do-Androids-Dream-of-Electric-Sheep?}).
	We want to infer more facts (viz. triplets) about {\it Blade-Runner} from the knowledge we already have,
		and answer questions such as ``Is Blade Runner science fiction?''
	If the knowledge base contains a triplet
		({\it Do-Androids-Dream-of-Electric-Sheep?}, {\it is-a}, {\it Science-fiction}),
		it should help us estimate that the answer is yes. 
	Figure~\ref{image:relation-example} illustrates this example schematically.

	There have been some attempts to obtain the embeddings for OOKB entities
		using external resources \cite{joint_align,joint_dis,Freebase_description}.
	Although these approaches may be useful,
		they require additional computation over large resources, which may not be always feasible.
	By contrast,
		we pursue a KBC model that exploits existing triplets in the knowledge base,
		without relying on external resources.
	Indeed, the {\it Blade Runner} example above suggests
		the possibility of inferring new facts about OOKB entities without the help of external resources.

	To solve the OOKB entity problem,
		we apply \emph{graph neural networks} (Graph-NNs) \cite{graphnnm,gate_gnn} to
	a {\em knowledge graph}, which is a graph obtained by regarding
		entities as nodes and triplets as edges.
	A Graph-NN is a neural network architecture defined on a graph structure
		and composed of two models called the {\em propagation model} and {\em output model}.
	The propagation model manages how information propagates between nodes in the graph.
	In the propagation model, we first obtain the embedding vectors of the neighborhood of a given node (entity) $e$ and then convert these vectors into a representation vector of $e$ using a pooling function such as the average.
	In other words, each node is embedded as a vector in continuous space, which is also used to calculate the vectors of the neighborhood nodes.
	This mechanism enables the vector for an OOKB entity to be composed from its neighborhood vectors at test time.
	The output model defines the task-oriented objective function over node vectors.
	This allows us to use an existing embedding-based KBC model as the output model.
	In this paper, we use TransE \cite{transE} as the output model, but we can adopt other embedding-based KBC methods.

	Our main contributions can be summarized as follows:
	\begin{itemize}
		\item We propose a new formulation of the problem of OOKB entities in KBC.
		\item We propose a Graph-NN suitable for the KBC task with OOKB entities.
		\item We verify the effectiveness of our models in both the standard and the ``OOKB'' entity settings.
	\end{itemize}


	\section{OOKB Entity Problem in Knowledge Base Completion}
\label{sec:ue}

	\subsection{Knowledge Graph}
		Let $\mathcal{E}$ be a set of \emph{entities} and $\mathcal{R}$ be a set of \emph{relations}.
		Define the \emph{fact}, or \emph{relation triplet}, to be a triplet of form $(h, r, t)$ where $h, t \in \mathcal{E}$ and $r \in \mathcal{R}$.
		Let $\KB_\gold \subset \mathcal{E} \times \mathcal{R} \times \mathcal{E}$  be the set of \emph{gold} facts,
			i.e., the set of all relation triplets that hold for pairs of entities in $\mathcal{E}$ and relations in $\mathcal{R}$.
		If a triplet is in $\KB_\gold$, we say it is a \emph{positive triplet}; otherwise, it is a \emph{negative triplet}.
		The goal of knowledge completion is to identify $\KB_\gold$,
			when only its proper subset, or an ``incomplete'' \emph{knowledge base}, $\KB \subset \KB_\gold$ is accessible.

	A knowledge base $\KB$ is often called \emph{knowledge graph}
		because each triplet in $\KB$ can be regarded as a (labeled) edge in a graph;
		i.e., the entities in a triplet correspond to the end nodes and the relation gives the label of the edge.

	\subsection{KBC: Triplet Classification}
		{\it Triplet classification} is a typical KBC task introduced by \cite{NTN}
			\footnote{A similar task had been around for general graphs under the name of \emph{link prediction}.}
			and has since been a standard benchmark for KBC methods \cite{transH,transD,transMani,lppTransX,neig}.

		In this task, existing knowledge base $\KB$ is assumed to be incomplete,
			in the sense that some of triplets that must be present in $\KB$ are missing;
			i.e., $\KB \ne \KB_\gold$.

		Let $\mathcal{H} = (\mathcal{E} \times \mathcal{R} \times \mathcal{E}) \backslash \KB$
			be the set of triplets not present in $\KB$.
		Because $\KB$ is incomplete,
			two cases are possible
			for each triplet $x \in \mathcal{H}$; 
			either $x$ is a positive triplet (i.e., $x \in \KB_\gold$),
			or $x$ is a negative triple (i.e., $x \not\in \KB_\gold$).
		For the former case, $x$ is not in $\KB$ only because of the incompleteness,
			and the knowledge base must be updated to contain $x$.
		We thus encounter the problem of determining
			which of the above two possible cases each triplet not present in $\KB$ falls into.
		This problem is called \emph{triplet classification}.

		Viewed as a machine learning problem,
			triplet classification is a classifier induction task in which
			$\mathcal{E}$ and $\mathcal{R}$ are given,
			and knowledge base $\KB$ forms the training set (with only positive examples),
			with $\mathcal{H}$ being the test set.
		The set $\mathcal{H}$ can be divided into the set of positive test examples
			$\mathcal{H} \cap \KB_\gold = \KB_\gold \backslash \KB$
			and the set of negative test examples $\mathcal{H} \backslash \KB_\gold$.

		In the standard triplet classification,
			$\mathcal{E}$ and $\mathcal{R}$ are limited to the entities and relations that appear in $\KB$.
		That is, $\mathcal{E}= \mathcal{E}(\KB)$ and $\mathcal{R}= \mathcal{R}(\KB)$,
			where
			$\mathcal{E}(\KB) = \{  h \mid (h,r,t) \in \KB \} \cup \{  t \mid (h,r,t) \in \KB \}$
			and
			$\mathcal{R}(\KB) = \{  r \mid (h,r,t) \in \KB \}$
			denote the entities and relations appearing in $\KB$, respectively.

	\subsection{OOKB Entity Problem}
		We now introduce a new task in KBC, called the OOKB entity problem.

		In addition to the knowledge base $\KB$ observed at training time,
			new triplets $\KB_\aux$ are given at test time,
			with $\mathcal{E}(\KB_\aux) \not \subset \mathcal{E}(\KB)$
			and $\mathcal{R}(\KB_\aux) \subseteq \mathcal{R}(\KB)$.
		Thus, $\KB_\aux$ contains new entities $\mathcal{E}_{\text{OOKB}} = \mathcal{E}(\KB_\aux) \backslash \mathcal{E}(\KB)$,
			but no new relations are involved.
		We call $\mathcal{E}_{\text{OOKB}}$ \emph{OOKB entities}.
		It is assumed that every triplet in $\KB_\aux$ contains
			exactly one OOKB entity from $\mathcal{E}_{\text{OOKB}}$ and one entity from $\mathcal{E}(\KB)$;
			that is, the additional triplets $\KB_\aux$ represent edges bridging $\mathcal{E}(\KB)$ and $\mathcal{E}_{\text{OOKB}}$
			in the combined knowledge graph $\KB \cup \KB_\aux$.
		In this setting, $\mathcal{E} = \mathcal{E}(\KB) \cup \mathcal{E}_{\text{OOKB}} \ne \mathcal{E}(\KB)$,
			and the task is to correctly identify missing relation triplets
			that involve the OOKB entities $\mathcal{E}_{\text{OOKB}}$.
		Because the embeddings for these entities are missing,
			they must be computed from those for entities in $\KB$.
		In other words, we want to design a model by which the information we already have in $\KB$
			can be transferred to OOKB entities $\mathcal{E}_{\text{OOKB}}$,
			with the help of the added knowledge $\KB_\aux$.


	\section{Proposed Model}
	\subsection{Graph-NNs}
		Graph-NNs are neural networks defined on a graph structure.
		Although there exist graph-NNs that encode an entire graph into a vector \cite{macro1,macro3},
			here we focus on the one that provides the means to encode nodes and edges into vectors,
			as this is more suitable for KBC.

		According to \cite{graphnnm,gate_gnn}, a graph-NN consists of two models, the propagation model and the output model.
		The propagation model determines how to propagate information between nodes in a graph.
		The output model defines an objective function according to given tasks using vector-represented nodes and edges.
		In this paper, we modify the propagation model to be suitable for knowledge graphs.
			For the output model, we use the embedding-based KBC model TransE \cite{transE}.

	\subsection{Propagation Model on a Knowledge Graph}
		Let $\KB$ be a knowledge graph, $e \in \mathcal{E}(\KB)$ be an entity, and $\mathbf{v}_e \in \mathbb{R}^d$ be the $d$-dimensional representation vector of $e$.
		Li et~al. define the propagation model by the following equation \cite{gate_gnn}:
		\begin{equation}
			\mathbf{v}_{e} = \!\!\!\!\!\!\!
			\sum_{(h,r,e) \in \mathcal{N}_\head(e)} \!\!\!\!\!\!\!\!\!
				T_\head(\mathbf{v}_{h}; h,r,e) 
			+ \!\!\!\!\!\!
			\sum_{(e,r,t) \in \mathcal{N}_\tail(e)} \!\!\!\!\!\!\!\!\!
				T_\tail(\mathbf{v}_{t}; e,r,t), 
			\label{eq:pre-pro}
		\end{equation}
		where head neighborhood $\mathcal{N}_\head$ and tail neighborhood $\mathcal{N}_\tail$ are $\mathcal{N}_\head(e) = \{(h,r,e) \mid (h, r, e) \in \mathcal{G} \}$ and $\mathcal{N}_\tail(e) = \{(e,r,t) \mid (e, r, t) \in \mathcal{G} \}$ in a knowledge graph $\mathcal G$, respectively.
		In addition, $T_\head, T_\tail : \mathbb{R}^d \times \mathcal{E}(\KB) \times \mathcal{R}(\KB) \times \mathcal{E}(\KB) \to \mathbb{R}^d$ are called \emph{transition functions} \cite{gate_gnn} and used to transform the vector of a neighbor node for its incorporation in the current vector $\mathbf{v}_e$, depending on the property of the edge between them.

		We generalize the propagation function using the following \emph{pooling function} $P$:
		\begin{align}
			S_\head(e)       & = \{ T_\head(\mathbf{v}_{h}; h,r,e) \mid (h,r,e) \in \mathcal{N}_h(e)\}, \label{eq:Sh} \\
			S_\tail(e)       & = \{ T_\tail(\mathbf{v}_{t}; e,r,t) \mid (e,r,t) \in \mathcal{N}_t(e)\}, \label{eq:St} \\
			\mathbf{v}_e & = P(S_\head(e) \cup S_\tail(e)), \label{eq:P(Sth)}
		\end{align}
		Here, $S_\head(e)$ contains the representation vectors of neighborhood $\mathcal{N}_\head(e)$, and $S_\tail(e)$ contains those for $\mathcal{N}_\tail(e)$.
		The difference between Eq.~\eqref{eq:pre-pro} and ours (Eqs.~\eqref{eq:Sh}--\eqref{eq:P(Sth)})
			is the use of the pooling function in place of the summation.
		The candidates for functions $T_\head$, $T_\tail$, and $P$ are described below.

		\paragraph{Transition Function}
			The aim of transition function $T$ (including both $T_\head$ and $T_\tail$) is to modify the vector of a neighbor node to reflect the relations between the current node and the neighbor.
			The examples of the transition function are listed here:
			\begin{align*}
				T(\mathbf{v})&=\mathbf{v},                     && \text{(identity)}             \\
				T(\mathbf{v})&=\tanh(\bm{A}\mathbf{v}),        && \text{(single $\tanh$ layer)} \\
				T(\mathbf{v})&=\mathrm{ReLU}(\bm{A}\mathbf{v}),  && \text{(single $\mathrm{ReLU}$ layer)}
			\end{align*}
			where $\bm{A} \in \mathbb R^{d \times d}$ is a matrix of model parameters and $\tanh$ and $\mbox{ReLU}$ are elementwise hyperbolic tangent and rectified linear unit functions.
			In addition, we can use other neural network techniques, such as batch-normalization \cite{BN}, residual connection \cite{resnet}, and long short term memory \cite{gate_gnn}.

			We can also make the transition function dependent on the relation between the current node (entity) and the neighbor, such as in the following:
			\begin{align*}
				T_\head(\mathbf{v}_{h};h,r,e)     &= \tanh(\bm{A}^\head_{(h,r,e)} \mathbf{v}_h), \\
				T_\tail(\mathbf{v}_{t};e,r,t)     &= \tanh(\bm{A}^\tail_{(e,r,t)} \mathbf{v}_t).
			\end{align*}
			Note that the parameter matrices are now defined individually for each combination of node $e$, the current neighbors ($h$ or $t$), and the relation $r$ between them.

			In the experiments of Section~\ref{sec:exp}, we use the following transition functions:
			\begin{align}
				T_\head(\mathbf{v}_{h}; h,r,e)     &= \mathop{\mathrm{ReLU}}(\mathop{\mathrm{BN}}(\bm{A}^\head_{r} \mathbf{v}_h)),\label{eq:our_head_transition} \\
				T_\tail(\mathbf{v}_{t}; e,r,t)     &= \mathop{\mathrm{ReLU}}(\mathop{\mathrm{BN}}(\bm{A}^\tail_{r} \mathbf{v}_t)),\label{eq:our_tail_transition}
			\end{align}
			where $\mathrm{BN}$ indicates batch normalization \cite{BN}.

		\paragraph{Pooling Function}
			\label{sec:pooling-function}
			Pooling function $P$ is a function that maps a set of vectors to a vector,
				i.e., $P: 2^{\mathbb{R}^d} \to \mathbb{R}^d$.
			Its objective is to extract shared aspects from a set of vectors.
			For $S = \{\mathbf{x}_i \in \mathbb{R}^d \}_{i=1}^{N}$,
				some simple pooling functions are as follows:
			\begin{align*}
				P(S) &=\sum_{i=1}^{N} \mathbf{x}_i,            &&  \text{(sum pooling)} \\
				P(S) &=\frac{1}{N}\sum_{i=1}^{N} \mathbf{x}_i, &&  \text{(average pooling)} \\
				P(S) &=\max(\{\mathbf{x}_i\}_{i=1}^N),         &&  \text{(max pooling)}
			\end{align*}
			where $\max$ is the elementwise max function.
			Sum pooling was used in \cite{graphnnm,gate_gnn}; see also Eq.~\eqref{eq:pre-pro}.

		\paragraph{Stacking and Unrolling Graph Neural Networks}
			As explained above, the propagation models decide how to propagate information from a node to its neighborhood.
			Applying this propagation model repeatedly, we can broadcast information of a node to farther nodes, i.e., each node can receive further information.
			Broadcasting can be implemented in one of two ways: stacking or unrolling.

			The unrolled Graph-NN is discussed in \cite{graphnnm,gate_gnn}.
			In the unrolled Graph-NN, the propagation model uses the same model parameters in every propagation.
			The propagation procedures are the same as described in Eqs.~\eqref{eq:Sh}--\eqref{eq:P(Sth)}.

			The stacked Graph-NN is constructed in a similar manner to the well-known stacking technique \cite{stack_tech}.
			In particular, the propagation process in the stacked Graph-NN uses different model parameters depending on time step $n$.
			The transition function at each time step $n$, indicated by superscript $n$, is as follows.
			\begin{equation*}
				  \mathbf{v}^{(n)}_{e} = \begin{cases}
					\mathbf{v}_{e},  & \text{if } n=0, \\
					P(S^{(n-1)}_\head(e) \cup S^{(n-1)}_\tail(e)), & \text{otherwise,}
				  \end{cases}
			\end{equation*}
			where
			\begin{align*}
			  S^{(n)}_\head(e) &= \{ T^{(n)}_\head(\mathbf{v}^{(n)}_{h}; h,r,e) \mid (h,r,e) \in \mathcal{N}_\head(e)\} \\
			  S^{(n)}_\tail(e) &= \{ T^{(n)}_\tail(\mathbf{v}^{(n)}_{t}; e,r,t) \mid (e,r,t) \in \mathcal{N}_\tail(e)\}.
			\end{align*}
			where $T^{(n)}_\head$ and $T^{(n)}_\tail$ are transition functions depending on head/tail and time.

	\subsection{Output Model: Score and Objective Functions}
		We use a TransE-based objective function as the output model.
		TransE \cite{transE} is one of the basic embedding-based models for KBC, and we use it for its simplicity and ease of training.
		Notice however that our architecture is not limited to TransE, and adopting other embedding-based models for the output model is equally straightforward.

		Below,
			we explain the score function of TransE and its commonly used \emph{pairwise-margin} objective functions.
		We then describe the modified objective function we used in our experiments, called the \emph{absolute-margin} objective.

		\paragraph{Score Function}
			The (implausibility) score function $f$ evaluates the implausibility of a triplet $(h, r, t)$;
			smaller scores indicate that the triplet is more likely to hold.
			In TransE, the score function is defined by $f(h,r,t) = \|\mathbf{v}_h+\mathbf{v}_r - \mathbf{v}_t \|$, where $\mathbf{v}_h$, $\mathbf{v}_r$, and $\mathbf{v}_t$ are the embedding vectors of the head, relation, and tail, respectively.
			This score function states that for a positive triplet $(h, r, t)$, the sum of the head and relation vectors $\mathbf{v}_h+\mathbf{v}_r$ must be close to the tail vector $\mathbf{v}_t$, i.e., $\mathbf{v}_h+\mathbf{v}_r \sim \mathbf{v}_t$.
			This score function is modified and extended in \cite{transH,transR,transD,transG}.
			As mentioned earlier, all these models can be used as our output model.

		\paragraph{Pairwise-Margin Objective Function}
			The objective (loss) function defines the quantity to be minimized through optimization.
			The following \emph{pairwise-margin} objective function is commonly used with KBC methods including TransE \cite{transE,transG}:
			\begin{eqnarray}
				\mathcal{L} = \sum_{i=1}^N \left[ \tau + f(h_i, r_i, t_i) - f(h'_i, r_i, t'_i) \right]_{+}
			\end{eqnarray}
			where $[x]_{+}$ is the hinge function $[x]_{+}=\max(0,x)$ and scalar $\tau \in \mathbb{R}$ is a threshold (called \emph{margin}),
			with $(h_i,r_i,t_i)$ denoting a positive triplet and $(h'_i,r_i,t'_i)$ denoting a negative triplet.
			This objective function requires score $f(h'_i, r_i, t'_i)$ to be greater than score $ f(h_i, r_i, t_i) $ by at least $\tau $.
			If the difference is smaller than $\tau$,
				then the optimization changes the parameters to meet the requirement.
			In contrast,
				if the difference is greater than $\tau$,
				the parameters are not updated.

			The pairwise-margin objective thus pays attention to the difference in scores between positive-negative triplet pairs.

		\paragraph{Absolute-Margin Objective Function}
			Instead of the pairwise-margin objective,
						in this paper, we employ the following objective function, which we call the {\em absolute-margin} objective.
			\begin{equation}
				\mathcal{L} = \sum_{i=1}^N f(h_i, r_i, t_i) + [\tau - f(h'_i, r_i, t'_i)]_{+}
					\label{eq:my-objective-function}
			\end{equation}
			where $\tau$ is a hyperparameter, again called the \emph{margin}.
			This objective function considers positive and negative triplets separately in the first and the second terms,
				not jointly as in the pairwise-margin objective.
			The scores for the positive triplets will be optimized towards zero,
			whereas the scores of the negative triplets are going to be at least $\tau$.
			This objective function not only is easy to optimize,
				but also obtained good results in our preliminary experiments.
			We thus used this objective function for the experiments in Section~\ref{sec:exp}.

\begin{table}[t]
	\caption{Specifications of the triplet classification datasets.
		  half of the validation and test sets are negative triplets, and these are included in the numbers of validation triplets and test triplets.
	}
	\label{table:stats_of_DS}
	\centering
	\small
	\begin{tabular}{l rr}
		  \toprule          & WordNet11 & Freebase13 \\
		  \midrule
		  Relations         & 11        & 13         \\
		  Entities          & 38,696    & 75,043     \\
		  Training triplets & 112,581   & 316,232    \\
		  Validation triplets & 5,218   & 11,816     \\
		  Test triplets     & 21,088    & 47,466     \\
		  \bottomrule
	\end{tabular}
\end{table}

\begin{table}[t]
	\caption{Result of the standard KBC experiment (without OOKB entities).
		  The figures represent accuracy.
		  Except for the proposed method, they are obtained from the respective papers.
		  Bold and underlined figures are the best and second best scores for each dataset, respectively.
	}
	\label{table:comparision_previous}
	\small
	\centering
	\begin{tabular}{l rr}
	\toprule
	Method			   & WordNet11	& Freebase13      \\ \midrule
	NTN \cite{NTN}  	   & 70.4	& 87.1	    \\
	TransE \cite{transE}	   & 75.9	& 81.5	    \\ 
	TransH \cite{transH}       & 78.8	& 83.3	    \\ 
	TransR \cite{transR}       & 85.9	& 82.5	    \\
	TransD \cite{transD}       & 86.4	& {\bf89.1} \\
	TransE-COMP \cite{trav}    & 80.3	& 87.6	    \\ 
	TranSparse \cite{Sparse}   & 86.8	& 88.2	    \\
	ManifoldE \cite{transMani} & \ul{87.5}  & 87.3	    \\
	TransG \cite{transG}	   & 87.4	& 87.3	    \\
	lppTransD \cite{lppTransX} & 86.2	& \ul{88.6} \\
	NMM \cite{neig}     & 86.8	& \ul{88.6} \\ 
	Proposed method		   & {\bf 87.8} & 81.6	    \\
	\bottomrule
	\end{tabular}
\end{table}

\begin{table*}[t]
	\caption{Number of entities and triplets in the the OOKB datasets.
		The numbers of triplets include negative triplets.
		}
	\label{table:stats_of_UE}
	\centering
	\small
	\begin{tabular}{l rrr rrr rrr}
		  \toprule
		  \multirow{2}{*}{}   & \multicolumn{3}{c}{Head} & \multicolumn{3}{c}{Tail} & \multicolumn{3}{c}{Both}                                     \\
		  \cmidrule(r){2-4} \cmidrule(lr){5-7} \cmidrule(l){8-10}
									  & 1,000                    & 3,000                    & 5,000  & 1,000  & 3,000  & 5,000  & 1,000  & 3,000  & 5,000  \\
		  \midrule
			  Training triplets   & 108,197                  & 99,963                   & 92,309 & 96,968 & 78,763 & 67,774 & 93,364 & 71,097 & 57,601 \\
			  Validation triplets & 4,613                    & 4,184                    & 3,845  & 3,999  & 3,122  & 2,601  & 3,799  & 2,759  & 2,166  \\
				  \midrule
			  OOKB entities       & 348                      & 1,034                    & 1,744  & 942    & 2,627  & 4,011  & 1,238  & 3,319  & 4,963  \\
			  Test triplets       & 994                      & 2,969                    & 4,919  & 986    & 2,880  & 4,603  & 960    & 2,708  & 4,196  \\
			  \midrule
			  Auxiliary entities  & 2,474                    & 6,791                    & 10,784 & 8,191  & 16,193 & 20,345 & 9,899  & 19,218 & 23,792 \\
			  Auxiliary triplets  & 4,352                    & 12,376                   & 19,625 & 15,277 & 31,770 & 40,584 & 18,638 & 38,285 & 48,425 \\
		  \bottomrule
	\end{tabular}
\end{table*}

\begin{table*}[t]
	\caption{Results of the OOKB experiment: accuracy of the simple baseline and proposed models.
	Bold and underlined figures are respectively the best and second best scores for each dataset.
	}
	\label{table:OOKB}
	\centering
		\small
	\begin{tabular}{lc rrr rrr rrr}
		  \toprule
		  \multicolumn{2}{c}{}           & \multicolumn{3}{c}{Head}          & \multicolumn{3}{c}{Tail}          & \multicolumn{3}{c}{Both}          \\
		  \cmidrule(r){3-5} \cmidrule(lr){6-8} \cmidrule(l){9-11}
			Method						& Pooling
				& 1,000      & 3,000      & 5,000
				& 1,000      & 3,000      & 5,000
				& 1,000      & 3,000      & 5,000      \\
		  \midrule

			\multirow{3}{*}{Baseline}	& sum
				& 54.6      & 52.5      & 52.0
				& 53.7      & 53.0      & 52.8
				& 54.0      & 52.7      & 53.2      \\

										& max
				& 58.1      & 56.3      & 56.4
				& 55.2      & 54.2      & 55.3
				& 56.8      & 56.8      & 56.4      \\

										& avg
				& 63.0      & 60.2      & 61.1
				& 63.8      & \ul{63.9} & \ul{63.0}
				& 65.3      & \ul{63.9} & \ul{64.8} \\

		  \midrule
		  \multirow{3}{*}{Proposed} 	& sum
				& 70.2  & 62.6  & 59.6
				& 64.6  & 56.5  & 55.0
				& 59.5  & 55.2  & 54.2  \\

										& max
				& \ul{80.3} & \ul{75.4} & \ul{72.7}
				& \ul{74.8} & 63.1      & 58.7
				& \ul{68.0} & 59.5      & 56.5      \\

										& avg
				& \bf 87.3  & \bf 84.3  & \bf 83.3
				& \bf 84.0  & \bf 75.2  & \bf 69.2
				& \bf 83.0  & \bf 73.3  & \bf 68.2  \\

		  \bottomrule
	\end{tabular}
\end{table*}


	\section{Experiments}
\label{sec:exp}
	\subsection{Implementation and Hyperparameters}
		We implemented our models using the neural network library Chainer \protect\url{http://chainer.org/}.
		All networks were trained by stochastic gradient descent with backpropagation;
		specifically, we used the Adam optimization method \cite{adam}.
		The step size of Adam was $\alpha_1 / (\alpha_2 \cdot k +1.0)$, where $k$ indicates the number of epochs performed,
				$\alpha_1=0.01$, and $\alpha_2=0.0001$.
		The mini-batch size was $5,000$ and the number of training epochs was $300$ in every experiment.
		Moreover, the dimension of the embedding space was $200$ in the standard triplet classification and $100$ in other settings.

		In the preliminary experiments,
			we tried several activation functions and pooling functions,
			and found the following hyperparameter settings
			on account of both computational time and performance.
		We used Eqs.~\eqref{eq:our_head_transition}--\eqref{eq:our_tail_transition} as transision functions in both the standard and OOKB settings.
		As the pooling function,
			we used the max pooling function in the standard triplet classification,
			and tried three pooling functions, max, sum, average, in the OOKB setting.
		The results of the preliminary experiments were reflected in our selection of the absolute-margin objective function over the pairwise-margin objective function
		as well as the margin value $\tau=300$ in the absolute-margin objective function (Eq.~\eqref{eq:my-objective-function}).
		The absolute-margin objective function converged faster than the pairwise-margin objective function.
		Because the task is a binary classification of triplets into positive (i.e., the relations that must be present in the knowledge base) and negative triplets (those that must not),
			we determined the threshold value for output scores between these classes using the validation data.

		To deal with the limited available computational resources (e.g., GPU memory),
			we sampled the neighbor entities randomly when an entity has too many of them.
		Indeed,
			there were some entities that appeared in a large number of the triplets,
			and thus had many neighbors;
			when the neighborhood size exceeded $64$, we randomly chose $64$ entities from the neighbors.

	\subsection{Standard Triplet Classification}
		We compared our model with the previous KBC models
			in the standard setting, in which no OOKB entities are involved.
		\paragraph{Datasets}
			We used WordNet11 and Freebase13 \cite{NTN} for evaluation.
			The data files were downloaded from \url{http://cs.stanford.edu/people/danqi/}.
			These datasets are subsets of two popular knowledge graphs, WordNet \cite{Wordnet} and Freebase \cite{Freebase}.
			The specifications on these datasets are shown in Table~\ref{table:stats_of_DS}.
			Both datasets contain training, validation, and test sets.
			The validation and test sets include positive and negative triplets.
			In contrast, the training set does not contain negative triplets.
			As usual with the case in which negative triplets are not available,
				\emph{corrupted} triplets are generated from positive triplets are used
				as a substitute for negative triplets.
			From a positive triplet $(h, r, t)$ in knowledge base $\KB$,
				a corrupted triplet is generated by substituting a random entity sampled from $\mathcal{E}(\KB)$ for $h$ or $t$.
			Specifically, to generate corrupted triplets, we used the ``Bernoulli'' trick, a technique
				also used in \cite{transH,transR,transD,Sparse}.

		\paragraph{Result}
			The results are shown in Table~\ref{table:comparision_previous}.
			Our model showed state-of-the-art performance on the WordNet11 dataset.
			On the Freebase13 dataset, it did not perform as well as the state-of-the-art KBC methods,
				although it was slightly better than TransE on which our model was built on.

	\subsection{OOKB Entity Experiment}
		\paragraph{Datasets}
			We processed the WordNet11 dataset to
				construct several datasets for our OOKB entity experiment.

			In total, nine datasets were constructed
				with different numbers and positions of OOKB entities
				sampled from the test set.
			The process consists of two steps:
				choosing OOKB entities and filtering and splitting triplets.
			\begin{enumerate}
				\item {\sl Choosing OOKB entities.}
				To choose the OOKB entities,
					we first selected $N=1,000$, $3,000$, and $5,000$ triplets
					from the WordNet11 test file.

				For each of these three sets,
					we chose the initial candidates for the OOKB entities (denoted by $\mathcal{I}$)
					in three different ways (thereby yielding nine datasets in total);
					these settings are called Head, Tail, and Both.
				In the Head setting,
					all head entities in the $N$ triplets are regarded as candidate OOKB entities.
				The Tail setting is similar, but with the tail entities regarded as candidates.
				In the Both setting, all entities appearing as either a head or tail are the candidates.

				The final OOKB entities are the entities $e \in \mathcal{I}$ that appear in a triplet $(e,r,e')$ or $(e',r,e)$ in the WordNet11 training set, with $e' \not\in \mathcal{I}$.
				Note that the entities $h,t \not\in \mathcal{I}$ are contained in the knowledge bases
				we already have and not in the OOKB entities.
				This last process filters out candidate OOKB entities
					that do not have any connection with the training entities.

				\item {\sl Filtering and splitting triplets.}
				Using the selected OOKB entities,
					the original training dataset was split into
					the training dataset and the auxiliary datasets for the OOKB entity problem.
				That is, triplets that did not contain the OOKB entities were placed in the OOKB training set,
					and triplets containing one OOKB entity and one non-OOKB entity were placed in the auxiliary set.
				Triplets that contained two OOKB entities were discarded.

				For the test triplets,
					we used the same first $N$ triplets in the WordNet11 test file that we used in Step~1,
					with the exception that the triplets that did not contain any OOKB entities were removed.
				For the validation triplets, we simply removed the triplets containing OOKB entities from the WordNet11 validation set.
			\end{enumerate}
			The details of the generated OOKB datasets are shown in Table~\ref{table:stats_of_UE}.
			We denote each of the nine datasets by
			$\{\text{Head},\text{Tail},\text{Both}\}$-$\{ \text{1,000}, \text{3,000}, \text{5,000} \}$,
				respectively,
				where the first part represents the position of OOKB entities
				and the second part represents the number of triplets used for generating the OOKB entities.

		\paragraph{Result}
			Using the nine datasets generated from WordNet11,
				we verified the effectiveness of our proposed model.

			We used the following simple method as the baseline in this experiment.
			Given an OOKB entity $u$,
				we first obtained the embedding vectors of the neighborhood
				(determined by the triplets in the auxiliary knowledge) using TransE,
				and then converted these vectors into the representation vector of $u$ using a pooling function: sum, max, or average.
			Note that because all the neighborhood entities of $u$ are in the training knowledge base,
				their vectors can be computed using standard KBC methods.
			We followed the original paper \cite{transE} of TransE for the hyperparameter and other settings.

			The results are shown in Table~\ref{table:OOKB}.
			The column labeled ``pooling'' indicates which pooling function was used.
			As the table shows, our model outperforms the baselines considerably.
			In particular, Graph-NN with average pooling outperforms the other methods on all datasets.
			Graph-NN with max pooling also shows good accuracy, but in several settings, such as $\text{Tail}$-$\text{3000}$, it was outperformed by the baseline model with average pooling.

	\subsection{Stacking and Unrolling Graph-NNs}
		Stacking and unrolling are important techniques because they enable us to broadcast node information to distant nodes.
		In this experiment, we illustrate the effect of stacking and unrolling in the standard triplet classification task.
		The dataset is WordNet11, used for standard triplet classification.

		\paragraph{Accuracy Comparison}
			Table~\ref{table:stack_unroll} shows the performance of the stacked and unrolled Graph-NNs.
			The parameter ``depth'' indicates how many times the propagation model is iteratively applied.
			Note that when $\text{depth}=1$, the two models reduce to the vanilla Graph-NN, for which the result was $87.8\%$, as also shown in Table~\ref{table:comparision_previous}.
			These results imply that the stacking and unrolling techniques do not improve performance.
			We believe this is because of the power of the embedding models, i.e., we can embed information about distant nodes into a continuous space.

\begin{table}[t]
	\caption{Accuracy of stacked and unrolled Graph-NNs. The depth indicates the number of times the propagation model is iteratively applied.}
	\label{table:stack_unroll}
	\small
	\centering
	\begin{tabular}{ccc}
	\toprule
	Depth		& Stacking		& Unrolling		\\ \midrule
	1			& 87.8			& 87.8			\\
	2			& 87.5			& 87.2			\\
	3			& 87.1			& 86.7			\\
	4			& 87.0			& 87.0			\\
	\bottomrule
	\end{tabular}
\end{table}


	\section{Conclusion}
	In this paper, we proposed a new KBC task in which entities unobserved at training time are involved.
	For this task, we proposed a Graph-NN tailored to KBC with OOKB entities.
	We conducted two triplet classification tasks to verify the effectiveness of our proposed model.
	In the OOKB entity problem,
		our model outperformed the baselines considerably.
	Our model also showed state-of-the-art performance on WordNet11 in the standard KBC setting.


	\section*{Acknowledgments}

We thank the anonymous reviewers for valuable comments.
This work was partially supported by JSPS Kakenhi Grant 15H02749.


	\bibliography{ref}

\begin{thebibliography}{}

\bibitem[\protect\citeauthoryear{Bollacker \bgroup \em et al.\egroup
  }{2008}]{Freebase}
Kurt Bollacker, Colin Evans, Praveen Paritosh, Tim Sturge, and Jamie Taylor.
\newblock Freebase: A collaboratively created graph database for structuring
  human knowledge.
\newblock In {\em Proceedings of the 2008 ACM SIGMOD International Conference
  on Management of Data}, pages 1247--1250, 2008.

\bibitem[\protect\citeauthoryear{Bordes \bgroup \em et al.\egroup
  }{2013}]{transE}
Antoine Bordes, Nicolas Usunier, Alberto Garcia-Duran, Jason Weston, and Oksana
  Yakhnenko.
\newblock Translating embeddings for modeling multi-relational data.
\newblock In {\em Advances in Neural Information Processing Systems 26}, pages
  2787--2795, 2013.

\bibitem[\protect\citeauthoryear{Cao \bgroup \em et al.\egroup }{2016}]{macro1}
Shaosheng Cao, Wei Lu, and Qiongkai Xu.
\newblock Deep neural networks for learning graph representations.
\newblock In {\em Proceedings of the 30th AAAI Conference on Artificial
  Intelligence}, 2016.

\bibitem[\protect\citeauthoryear{Defferrard \bgroup \em et al.\egroup
  }{2016}]{macro3}
Micha\"{e}l Defferrard, Xavier Bresson, and Pierre Vandergheynst.
\newblock Convolutional neural networks on graphs with fast localized spectral
  filtering.
\newblock In {\em Advances in Neural Information Processing Systems 29}, pages
  3844--3852, 2016.

\bibitem[\protect\citeauthoryear{Fang \bgroup \em et al.\egroup
  }{2016}]{joint_dis}
Wei Fang, Jianwen Zhang, Dilin Wang, Zheng Chen, and Ming Li.
\newblock Entity disambiguation by knowledge and text jointly embedding.
\newblock In {\em Proceedings of The 20th SIGNLL Conference on Computational
  Natural Language Learning}, pages 260--269, 2016.

\bibitem[\protect\citeauthoryear{Guu \bgroup \em et al.\egroup }{2015}]{trav}
Kelvin Guu, John Miller, and Percy Liang.
\newblock Traversing knowledge graphs in vector space.
\newblock In {\em Proceedings of the 2015 Conference on Empirical Methods in
  Natural Language Processing}, pages 318--327, 2015.

\bibitem[\protect\citeauthoryear{He \bgroup \em et al.\egroup }{2016}]{resnet}
Kaiming He, Xiangyu Zhang, Shaoqing Ren, and Jian Sun.
\newblock Deep residual learning for image recognition.
\newblock In {\em Proceedings of the IEEE Conference on Computer Vision and
  Pattern Recognition}, 2016.

\bibitem[\protect\citeauthoryear{Ioffe and Szegedy}{2015}]{BN}
Sergey Ioffe and Christian Szegedy.
\newblock Batch normalization: Accelerating deep network training by reducing
  internal covariate shift.
\newblock In {\em Proceedings of the 32nd International Conference on Machine
  Learning}, pages 448--456, 2015.

\bibitem[\protect\citeauthoryear{Ji \bgroup \em et al.\egroup }{2015}]{transD}
Guoliang Ji, Shizhu He, Liheng Xu, Kang Liu, and Jun Zhao.
\newblock Knowledge graph embedding via dynamic mapping matrix.
\newblock In {\em Proceedings of the 53rd Annual Meeting of the Association for
  Computational Linguistics and the 7th International Joint Conference on
  Natural Language Processing}, volume 1: Long Papers, pages 687--696, 2015.

\bibitem[\protect\citeauthoryear{Ji \bgroup \em et al.\egroup }{2016}]{Sparse}
Guoliang Ji, Kang Liu, Shizhu He, and Jun Zhao.
\newblock Knowledge graph completion with adaptive sparse transfer matrix.
\newblock In {\em Proceedings of the 30th AAAI Conference on Artificial
  Intelligence}, 2016.

\bibitem[\protect\citeauthoryear{Kingma and Ba}{2014}]{adam}
Diederik~P. Kingma and Jimmy Ba.
\newblock Adam: {A} method for stochastic optimization.
\newblock {\em CoRR}, abs/1412.6980, 2014.

\bibitem[\protect\citeauthoryear{Li \bgroup \em et al.\egroup
  }{2015}]{gate_gnn}
Yujia Li, Daniel Tarlow, Marc Brockschmidt, and Richard~S. Zemel.
\newblock Gated graph sequence neural networks.
\newblock {\em CoRR}, abs/1511.05493, 2015.

\bibitem[\protect\citeauthoryear{Lin \bgroup \em et al.\egroup }{2015}]{transR}
Yankai Lin, Zhiyuan Liu, Maosong Sun, Yang Liu, and Xuan Zhu.
\newblock Learning entity and relation embeddings for knowledge graph
  completion.
\newblock In {\em Proceedings of the 29th AAAI Conference on Artificial
  Intelligence}, 2015.

\bibitem[\protect\citeauthoryear{Miller}{1995}]{Wordnet}
George~A. Miller.
\newblock Wordnet: A lexical database for {E}nglish.
\newblock {\em Communication of the ACM}, 38(11):39--41, 1995.

\bibitem[\protect\citeauthoryear{Nguyen \bgroup \em et al.\egroup
  }{2016}]{neig}
Dat~Quoc Nguyen, Kairit Sirts, Lizhen Qu, and Mark Johnson.
\newblock Neighborhood mixture model for knowledge base completion.
\newblock In {\em Proceedings of The 20th SIGNLL Conference on Computational
  Natural Language Learning}, pages 40--50, 2016.

\bibitem[\protect\citeauthoryear{Nickel \bgroup \em et al.\egroup
  }{2016}]{review}
M.~Nickel, K.~Murphy, V.~Tresp, and E.~Gabrilovich.
\newblock A review of relational machine learning for knowledge graphs.
\newblock {\em Proceedings of the IEEE}, 104:11--33, 2016.

\bibitem[\protect\citeauthoryear{Scarselli \bgroup \em et al.\egroup
  }{2009}]{graphnnm}
F.~Scarselli, M.~Gori, A.~C. Tsoi, M.~Hagenbuchner, and G.~Monfardini.
\newblock The graph neural network model.
\newblock {\em IEEE Transactions on Neural Networks}, 20:61--80, 2009.

\bibitem[\protect\citeauthoryear{Socher \bgroup \em et al.\egroup }{2013}]{NTN}
Richard Socher, Danqi Chen, Christopher~D Manning, and Andrew Ng.
\newblock Reasoning with neural tensor networks for knowledge base completion.
\newblock In {\em Advances in Neural Information Processing Systems 26}, pages
  926--934, 2013.

\bibitem[\protect\citeauthoryear{Vincent \bgroup \em et al.\egroup
  }{2010}]{stack_tech}
Pascal Vincent, Hugo Larochelle, Isabelle Lajoie, Yoshua Bengio, and
  Pierre-Antoine Manzagol.
\newblock Stacked denoising autoencoders: Learning useful representations in a
  deep network with a local denoising criterion.
\newblock {\em Journal of Machine Learning Research}, 11:3371--3408, 2010.

\bibitem[\protect\citeauthoryear{Wang \bgroup \em et al.\egroup
  }{2014a}]{joint_align}
Zhen Wang, Jianwen Zhang, Jianlin Feng, and Zheng Chen.
\newblock Knowledge graph and text jointly embedding.
\newblock In {\em Proceedings of the 2014 Conference on Empirical Methods in
  Natural Language Processing}, pages 1591--1601, 2014.

\bibitem[\protect\citeauthoryear{Wang \bgroup \em et al.\egroup
  }{2014b}]{transH}
Zhen Wang, Jianwen Zhang, Jianlin Feng, and Zheng Chen.
\newblock Knowledge graph embedding by translating on hyperplanes.
\newblock In {\em Proceedings of the 28th {AAAI} Conference on Artificial
  Intelligence}, 2014.

\bibitem[\protect\citeauthoryear{Xiao \bgroup \em et al.\egroup
  }{2016a}]{transMani}
Han Xiao, Minlie Huang, and Xiaoyan Zhu.
\newblock From one point to a manifold: Knowledge graph embedding for precise
  link prediction.
\newblock In {\em Proceedings of the 25th International Joint Conference on
  Artificial Intelligence}, pages 1315--1321, 2016.

\bibitem[\protect\citeauthoryear{Xiao \bgroup \em et al.\egroup
  }{2016b}]{transG}
Han Xiao, Minlie Huang, and Xiaoyan Zhu.
\newblock Trans{G}: A generative model for knowledge graph embedding.
\newblock In {\em Proceedings of the 54th Annual Meeting of the Association for
  Computational Linguistics}, volume 1: Long Papers, pages 2316--2325, 2016.

\bibitem[\protect\citeauthoryear{Yoon \bgroup \em et al.\egroup
  }{2016}]{lppTransX}
Hee-Geun Yoon, Hyun-Je Song, Seong-Bae Park, and Se-Young Park.
\newblock A translation-based knowledge graph embedding preserving logical
  property of relations.
\newblock In {\em Proceedings of the 2016 Conference of the North American
  Chapter of the Association for Computational Linguistics: Human Language
  Technologies}, pages 907--916, 2016.

\bibitem[\protect\citeauthoryear{Zhong \bgroup \em et al.\egroup
  }{2015}]{Freebase_description}
Huaping Zhong, Jianwen Zhang, Zhen Wang, Hai Wan, and Zheng Chen.
\newblock Aligning knowledge and text embeddings by entity descriptions.
\newblock In {\em Proceedings of the 2015 Conference on Empirical Methods in
  Natural Language Processing}, pages 267--272, 2015.

\end{thebibliography}
	\bibliographystyle{named}
\end{document}